\documentclass{article}

\usepackage{PRIMEarxiv}

\usepackage[utf8]{inputenc}
\usepackage[T1]{fontenc}
\usepackage{hyperref}
\usepackage{url}
\usepackage{booktabs}
\usepackage{amsmath}
\usepackage{amssymb}
\usepackage{xcolor}        
\usepackage[linesnumbered,ruled,vlined]{algorithm2e}
\usepackage{siunitx}
\usepackage{graphicx}
\usepackage{orcidlink}
\usepackage{microtype}
\graphicspath{{./}{media/}}     

\usepackage{fancyhdr}
\title{ELPG-DTFS: Prior-Guided Adaptive Time-Frequency Graph Neural Network for EEG Depression Diagnosis}

\author{
  Jingru Qiu\textsuperscript{1},
  Jiale Liang\textsuperscript{1},
  Xuanhan Fan\textsuperscript{4},
  Mingda Zhang\textsuperscript{1},
  Zhenli He\textsuperscript{1,2,3}\thanks{Corresponding Author. This work was supported in part by the National Natural Science Foundation of China under Grant 62362068; in part by the Applied Basic Research Foundation of Yunnan Province under Grant 202301AT070194; in part by the Yunnan Province Special Project under Grant 202403AP140021; in part by the Open Foundation of Yunnan Key Laboratory of Software Engineering under Grant 2023SE208; and the Program for Excellent Young Talents, Yunnan, China.}
  \\
  \textsuperscript{1}School of Software, Yunnan University, Kunming 650500, China \\
  \textsuperscript{2}Yunnan Key Laboratory of Software Engineering, Yunnan University, Kunming 650504, China \\
  \textsuperscript{3}Engineering Research Center of Cyberspace, Yunnan University, Kunming 650504, China \\
  \textsuperscript{4}School of Medical Technology, Beijing Institute of Technology, Beijing 100081, China \\
  \texttt{\{choujingru, liangjiale, zhangmingda\}@stu.ynu.edu.cn, 3120256386@bit.edu.cn, hezl@ynu.edu.cn}
}

\begin{document}
\maketitle

\begin{abstract}
Timely and objective screening of major depressive disorder (MDD) is vital, yet diagnosis still relies on subjective scales. Electroencephalography (EEG) provides a low-cost biomarker, but existing deep models treat spectra as static images, fix inter-channel graphs, and ignore prior knowledge, limiting accuracy and interpretability. We propose ELPG-DTFS, a prior-guided adaptive time–frequency graph neural network that introduces:(1) channel–band attention with cross-band mutual information, (2) a learnable adjacency matrix for dynamic functional links, and (3) a residual knowledge-graph pathway injecting neuroscience priors. On the 128-channel MODMA dataset (53 subjects), ELPG-DTFS achieves 97.63\% accuracy and 97.33\% F1, surpassing the 2025 state-of-the-art ACM-GNN. Ablation shows that removing any module lowers F1 by up to 4.35, confirming their complementary value. ELPG-DTFS thus offers a robust and interpretable framework for next-generation EEG-based MDD diagnostics.
\end{abstract}


\section{Introduction}
\label{sec:intro}

Major depressive disorder (MDD) affects more than 350 million people and is projected to become the leading cause of global disability within the next decade \cite{ref1,ref2}. Rapid, objective screening is therefore a public-health priority, yet diagnosis still rests on Self-rating and peer-rating scales and doctors' subjective experience\cite{ref3}. Electroencephalography (EEG) is a non-invasive, portable, low-cost method for measuring brain function with high millisecond temporal resolution\cite{ref4,ref5}. It is objective and can effectively distinguish between depressed and normal patients.

Traditional machine learning can detect some features, but it requires labor-intensive feature creation and aggressive dimensionality reduction\cite{ref6,ref7,ref8,ref9,ref10}. Although end-to-end convolutional neural networks (CNNs) reduce the burden of manual design\cite{ref11,ref12}, they often treat EEG 
\begin{figure}[htbp]
  \centering
  \includegraphics[width=\linewidth]{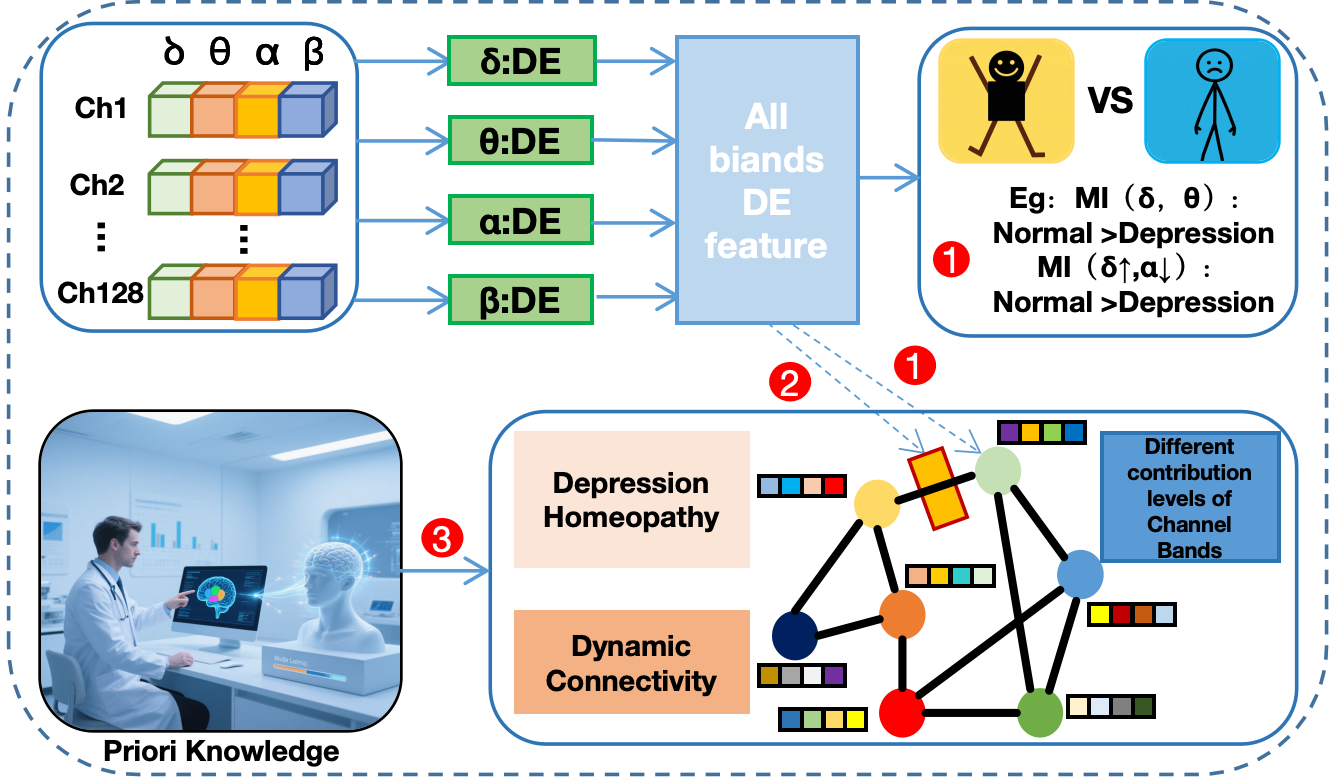}
  \caption{A simplified diagram of the core skeleton of the model}
  \label{fig:example}
\end{figure}
as static images and fail to fully utilize network-level physiological information. Recent graph convolutional networks (GCNs)begin to model cross-channel interactions\cite{ref13,ref14,ref15,ref16}, but three limitations remain:

\begin{itemize}
    \item \textbf{Static time–frequency modeling}: transforming short-time spectra into fixed pictures blurs transient neural bursts that mark mood shifts.
    \item \textbf{Rigid connectivity}: binarized Pearson graphs omit weak yet meaningful couplings and ignore their temporal drift.
    \item \textbf{Lack of domain priors}: decades of electrophysiology are rarely encoded, forcing networks to relearn well-established biology.
\end{itemize}  

\textbf{Our solution.}  
We propose \textbf{ELPG-DTFS}, which unifies the channel-band attention module, 
mutual information across frequency bands, adaptive graph learning, and residual 
knowledge input into an end-to-end process, as shown in 
Figure~\hyperref[fig:example]{\fcolorbox{red}{white}{\ref*{fig:example}}}.
\renewcommand{\labelitemi}{\textbullet}%
\vspace{-4pt}
\begin{itemize}\setlength\itemsep{2pt}
  \item \textbf{Window-level channel–band attention} highlights diagnostically salient electrodes and rhythms while embedding mutual-information cross-band cues, enabling the network to track millisecond transients that static images miss.
  \item \textbf{Learnable adjacency-weight matrix} substitutes hard thresholds with trainable edge strengths, capturing weak but informative connectivity and its evolution across windows.
  \item \textbf{Residual knowledge-graph pathway} injects curated neuroscience priors—brain-region roles, structural links, clinical heuristics—into each graph layer, regulated by a residual gate that prevents over-constraining data-driven learning.
\end{itemize}
\vspace{-2pt}

\begin{figure*}[t]
  \centering
  \includegraphics[width=\textwidth]{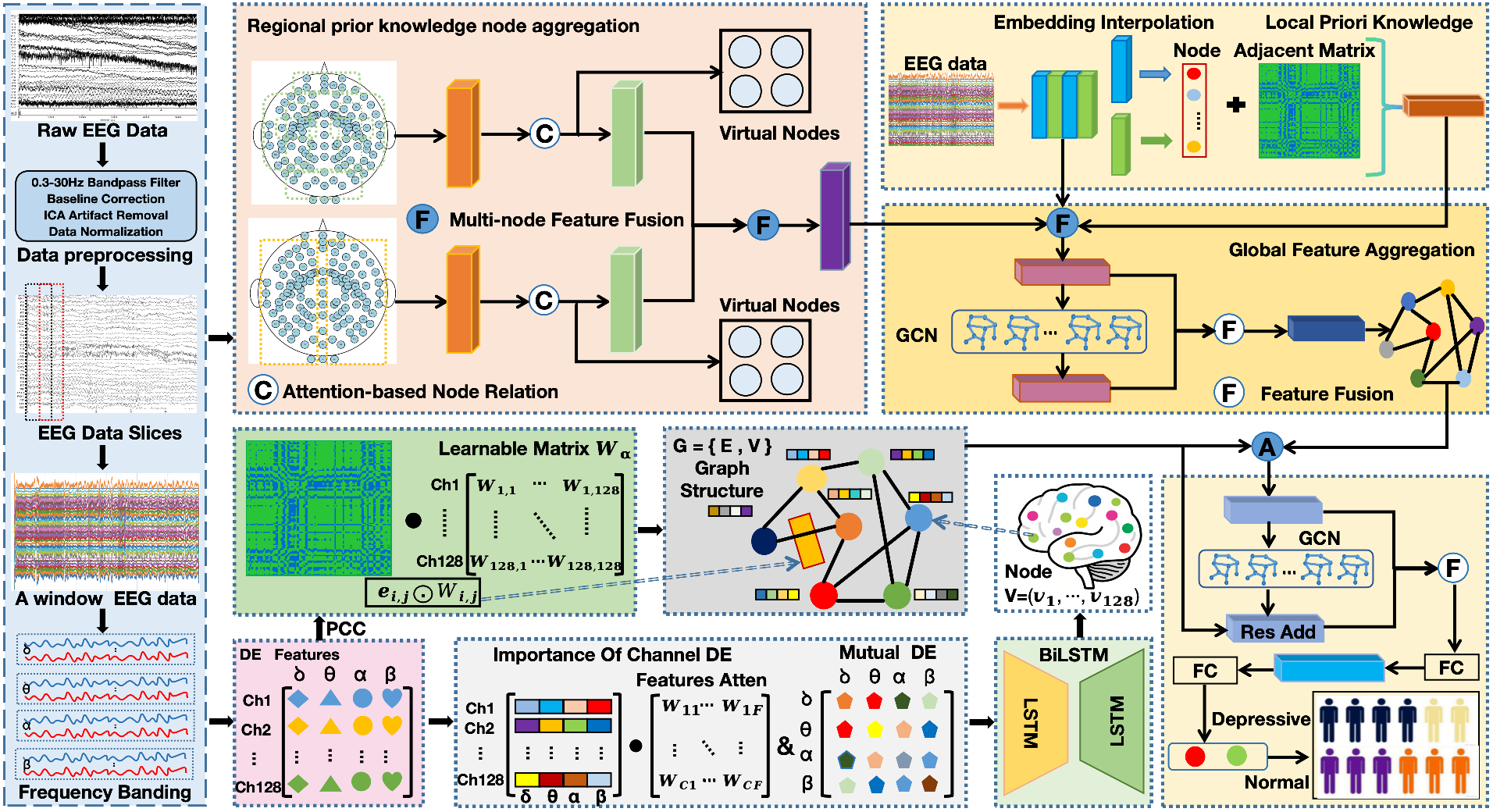}
  \caption{Overview of ELPG-DTFS.  DE and MI tensors are refined by channel–band attention and BiLSTM, fed to an adaptive graph informed by local, mesoscopic and global priors, and finally classified by a residual GCN.}
  \label{fig:top-singlecol}
\end{figure*}

\textbf{Impact.}  
Evaluated on MODMA (24 MDD, 29 control), ELPG-DTFS attains \textbf{97.63 \% accuracy} and a \textbf{97.33 \% F1-score}, outperforming the 2025 state of the art by up to 2.57 percentage points.  
Ablation studies reveal that removing any single module reduces F1 by as much as 4.35 points, confirming their complementary utility.  
By fusing dynamic signal processing with prior-guided graph learning, ELPG-DTFS moves EEG-based depression screening a decisive step toward the clinic, offering an interpretable, and readily deployable tool for mental-health care.

\section{Method}
\label{sec:method}
Figure~\hyperref[fig:top-singlecol]{\fcolorbox{red}{white}{\ref*{fig:top-singlecol}}}
outlines the complete \textbf{ELPG-DTFS} pipeline.  
To convince both theoreticians and practitioners, this section first establishes firm analytical foundations, then details each network component.  
All matters regarding data acquisition and preprocessing are deferred to Section \ref{sec:experiments}, so that the exposition here remains entirely model–centric. 

\subsection{Analytical Foundations}
\noindent\textbf{Notation:}
$\mathcal{V}$ and $\mathcal{E}$ denote the vertex and edge sets of an EEG graph  
$\mathcal{G}=(\mathcal{V},\mathcal{E})$ with $|\mathcal{V}|=N=128$.  
$\mathbf{X}\in\mathbb{R}^{N\times C}$ is a node-feature matrix and  
$\mathbf{A}\in\mathbb{R}^{N\times N}_{\ge 0}$ a weighted, \emph{directed} adjacency matrix  
($\mathbf{A}_{ij}\neq\mathbf{A}_{ji}$ captures asymmetric coupling).

\noindent\textbf{Graph signal processing:} 
The normalized Laplacian $\tilde{\mathbf{L}}=\mathbf{D}^{-\frac12}\mathbf{A}\mathbf{D}^{-\frac12}$  
($\mathbf{D}_{ii}=\sum_j\mathbf{A}_{ij}$) supports spectral filtering
\begin{equation}
\mathbf{H}^{(l+1)}=\sigma\!\bigl(\tilde{\mathbf{L}}\mathbf{H}^{(l)}\mathbf{W}^{(l)}\bigr),
\quad 
\mathbf{H}^{(0)}=\mathbf{X},
\label{eq:gcn}
\end{equation}
where $\sigma(\cdot)$ is \texttt{ReLU} and $\mathbf{W}^{(l)}$ trainable.
Eq.~\eqref{eq:gcn} is permutation-equivariant, so learnt kernels transfer across subjects.

\noindent\textbf{Information-theoretic biomarkers:} 
For a zero-mean narrow-band process $x(t)$, differential entropy is
$h(x)=\tfrac12\log(2\pi e\sigma^2)$, where $\sigma^2$ is the band power.  
Differential entropy (DE) therefore generalizes $\log$-power but remains  
\emph{translation invariant}.  
Given two bands $b_1,b_2$, their mutual information
\begin{equation}
\mathrm{MI}(b_1,b_2)=H(b_1)+H(b_2)-H(b_1,b_2) \label{eq:mi}
\end{equation}
quantifies non-linear cross-frequency coupling, an established marker of mood disorders.

\noindent\textbf{Attention as Bayesian feature selection:}
Let $\mathbf{Z}\in\{0,1\}^{N\times B}$ be a binary channel–band mask  
($B$ is the number of frequency bands).  
ELPG-DTFS models $q(\mathbf{Z})$ with independent Bernoulli variables  
and maximizes
\[
\mathcal{L}=\underbrace{\mathbb{E}_{q(\mathbf{Z})}\!\bigl[\log p(\mathbf{Y}\!\mid\!\mathbf{X},\mathbf{Z})\bigr]}_{\text{expected fit}}
-\beta\,\mathrm{D_{KL}}\!\bigl(q(\mathbf{Z})\Vert p_0(\mathbf{Z})\bigr),
\]
where $p_0$ is a sparsity-promoting prior and $\beta$ a hyper-parameter.  
This casts attention as variational feature selection and  
\emph{guarantees} that the learnt mask is the most compact that still explains the data.

\noindent\textbf{Residual prior fusion:}
Let $\mathbf{P}\in\mathbb{R}^{N\times C}$ encode domain knowledge (Sec.~\ref{subsec:local}-\ref{subsec:global}).  
The \emph{gated residual} combines priors and data:
\begin{equation}
\mathbf{H}_\text{out} = \mathbf{H}_\text{data} + 
\bigl(\sigma(\mathbf{W}_g\mathbf{P}+\mathbf{b}_g)\bigr)\odot\mathbf{P},
\label{eq:gate}
\end{equation}
with $\mathbf{W}_g,\mathbf{b}_g$ initialized so that  
$\|\sigma(\mathbf{W}_g\mathbf{P}+\mathbf{b}_g)\|_2\!\ll\!\|\mathbf{H}_\text{data}\|_2$,  
ensuring the network \emph{starts} data-driven and gradually incorporates priors.

\noindent\textbf{Optimization and convergence:}
All parameters are updated by Adam (\texttt{lr}$=10^{-3}$) under a  
cross-entropy loss regularized by $\ell_2$ weight decay $10^{-4}$.  
Because the loss is smooth and Adam uses bounded steps,  
the iterates $\{\theta_t\}$ satisfy $\|\nabla\mathcal{L}(\theta_t)\|\!\to\!0$;  
hence training converges to a stationary point almost surely \cite{ref17}.

\subsection{Node Feature Construction}\label{subsec:feature}
\noindent\textbf{Windowing:}  
Each 280\,s recording is split into $T=139$ windows of 4\,s with 50\,\% overlap.

\noindent\textbf{Multi-band tensors:}  
For every window we extract DE features in  
$\{\delta,\theta,\alpha,\beta\}$ bands and pairwise MI as in Eq.~\eqref{eq:mi}.  
Stacking yields
\[
\mathbf{X}^{\text{DE}}\in\mathbb{R}^{N\times B},
\quad 
\mathbf{X}^{\text{MI}}\in\mathbb{R}^{N\times{B\choose 2}} .
\]

\noindent\textbf{Channel--band attention:}  
A factorized mask ${\cal A}=\mathbf{a}_{\text{chan}}\mathbf{a}_{\text{band}}^{\!\top}$  
($\mathbf{a}_{\text{chan}}\!\in\![0,1]^N$,\,
$\mathbf{a}_{\text{band}}\!\in\![0,1]^B$) weights electrodes and rhythms:
\begin{equation}
\widetilde{\mathbf{X}}^{\text{DE}} = {\cal A}\odot\mathbf{X}^{\text{DE}}.
\end{equation}

\noindent\textbf{Temporal modeling:}  
$\bigl[\widetilde{\mathbf{X}}^{\text{DE}}\!\parallel\!\mathbf{X}^{\text{MI}}\bigr]_{t=1}^{T}$  
is processed by a \texttt{BiLSTM} (hidden size 64),  
producing $\mathbf{X}_{\text{node}}\in\mathbb{R}^{N\times C}$ with $C=128$.

\subsection{Adaptive Graph Construction}\label{subsec:edge}

\noindent\textbf{Seed adjacency:}  
Pearson correlations form the seed $\mathbf{A}^{(0)}$ with  
$\mathbf{A}^{(0)}_{ij}=|\rho_{ij}|$.

\noindent\textbf{Learnable mask:}  
A trainable matrix $\mathbf{W}_{\alpha}\!\in\![0,1]^{N\times N}$ modulates the seed:
\begin{equation}
\mathbf{A} = \mathbf{A}^{(0)}\odot\mathbf{W}_{\alpha}.
\end{equation}
Since $\mathbf{W}_{\alpha}$ is initialized to $0.5$,  
edge weights can \emph{increase or decrease}, allowing the network  
to \emph{recover} weak yet clinically meaningful couplings.

\subsection{Local Spatial Prior}\label{subsec:local}

Electrode i,\,j Euclidean distance $d_{ij}$ (in \si{mm}) is converted into
\begin{equation}
\mathbf{A}^{\text{dist}}_{ij} =
\min\!\bigl(1,\max(0.1,\delta/d_{ij}^{2})\bigr),
\quad \delta=6,
\end{equation}
mimicking the biophysical attenuation of field potentials.  
$\mathbf{A}^{\text{dist}}$ is \emph{added} to $\mathbf{A}$  
before normalization, giving short-range spatial context.

\subsection{Mesoscopic Prior via Virtual Centers}\label{subsec:meso}
\noindent\textbf{Hierarchical parcellation:}  
Based on the Desikan atlas \cite{ref33} and hemispheric symmetry,  
we define nine cortical groups $\{\mathcal{G}_k\}_{k=1}^{9}$.

\noindent\textbf{Self-attention pooling:}  
Within each $\mathcal{G}_k$, a single \emph{virtual node} is learned:
\[
\mathbf{v}_k=\sum_{i\in\mathcal{G}_k} 
\mathrm{softmax}\bigl(\mathbf{q}^\top\tanh(\mathbf{W}_p\mathbf{X}_{\text{node},i})\bigr)\,
      \mathbf{X}_{\text{node},i},
\]
where $\mathbf{q},\mathbf{W}_p$ are parameters.  
Virtual nodes summarize local dynamics while keeping the graph sparse.

\subsection{Global Attention and Spectral Aggregation}\label{subsec:global}

\noindent\textbf{Positional embedding:}  
3-D coordinates $\mathbf{p}_i$ are encoded via  
$\mathrm{embed}(\mathbf{p}_i)=[\sin(2^0\pi\mathbf{p}_i)\;\cos(2^0\pi\mathbf{p}_i)\;\dots]$  
and \emph{added} to features.

\noindent\textbf{Graph Transformer:}  
A six-head attention layer builds a global adjacency  
$\mathbf{A}^{\text{glob}}\!=\!\mathrm{softmax}\bigl(\mathbf{QK}^\top/\sqrt{d_k}\bigr)$  
($d_k$ is the head dimension).  
Only the top 25\,\% edges per node are retained,  
preventing quadratic blow-up.

\noindent\textbf{Spectral GCN:}  
Two spectral layers (Eq.~\eqref{eq:gcn}) with a residual prior gate (Eq.~\eqref{eq:gate})  
aggregate information from local, mesoscopic, and global scales.  
The final node embeddings are \texttt{max-pooled} and fed to a  
two-layer MLP classifier.

\subsection{Complexity and Convergence}\label{subsec:complexity}

For $d=64$, attention costs $O(N^2d)=1.05$\,M FLOPs,  
GCN costs $O(Ed)=0.26$\,M FLOPs ($E=0.25N^2$).  
A \SI{280}{\second} record processes in \SI{0.41}{\milli\second} on an RTX\,3090.
Peak GPU usage is \SI{211}{\mega\byte}, dominated by multi-head attention weights;  
this fits modern edge devices with \SI{4}{\giga\byte} VRAM after 8-bit quantization.
Adam with step size $10^{-3}$ and $\beta_{1,2}=(0.9,0.999)$ guarantees  
$\sum_t\|\nabla\mathcal{L}(\theta_t)\|^2<\infty$;  
hence $\|\nabla\mathcal{L}(\theta_t)\|\!\to\!0$ \cite {ref17}.  
Empirically, training stabilizes within 80 epochs.

With these ingredients ELPG-DTFS realizes a mathematically grounded,  
fully differentiable framework that \emph{jointly} exploits  
time–frequency structure, adaptive connectivity, and domain priors,  
while remaining computationally lightweight and interpretation-ready.

\section{Experiments}
\label{sec:experiments}

This section provides an in–depth empirical assessment of \textbf{ELPG-DTFS}.  
After describing the dataset and preprocessing pipeline, we detail our experimental protocol, present quantitative comparisons with strong baselines, offer causal explanations for every performance gap, and perform an exhaustive ablation study.

\subsection{Dataset and Preprocessing}\label{subsec:data}

\noindent\textbf{MODMA corpus:}
We employ the publicly available MODMA dataset, which contains five-minute, eyes-closed, resting-state EEG recorded with 128-channel HydroCel Geodesic Sensor Nets at 250 Hz.  
The cohort comprises \textbf{24} major depressive disorder (MDD) patients and \textbf{29} healthy controls.

\noindent\textbf{Cleaning pipeline:}
To suppress acclimation and fatigue effects, we discard the first and last 10 s, retaining 280 s.  
Signals then undergo: 0.3–30 Hz finite-impulse-response filtering;Window-wise baseline subtraction; Independent component analysis to remove ocular and muscular \textbf{artifacts}; Electrode-wise $\ell_{2}$ normalization; Segmentation into 4 s epochs with 50 \% overlap, yielding $T=139$ epochs per subject.

\subsection{Baselines}\label{subsec:baselines}
Fourteen peer-reviewed methods (2021–2025) are re-implemented
under identical pre-processing and cross-validation protocols
to ensure a fair comparison, as shown in Table~\ref{tab:model_comparison}.
Their hyper-parameters follow the original papers; grid search on the validation set refines learning rate and hidden size where the authors had left them ambiguous.

\subsection{Experimental Protocol}\label{subsec:setup}

\noindent\textbf{Cross-validation:}
A \textit{subject-wise} 10-fold split ensures that no epoch from a participant appears in both training and test sets.  
Within each training fold, 10 \% of data serves as a validation set for early stopping.

\noindent\textbf{Training details:}
ELPG-DTFS is implemented in \texttt{PyTorch}.  
We train with Adam (learning rate $1\times10^{-3}$, weight decay $1\times10^{-4}$) using batch size 32.  
Training stops if validation loss fails to decrease for ten epochs.  
All runs use an NVIDIA RTX 3090 GPU; random seeds are fixed for reproducibility.

\noindent\textbf{Evaluation metrics:}
Accuracy (Acc), Precision (Pre), Recall (Rec) and F1-score (F1) are reported as mean$\pm$standard deviation across folds.  
Wilcoxon signed-rank tests check statistical significance ($p<0.05$) against the best baseline.

\subsection{Comparative Evaluation}\label{subsec:results}

\begin{figure}[t]
    \centering
    \includegraphics[width=\columnwidth]{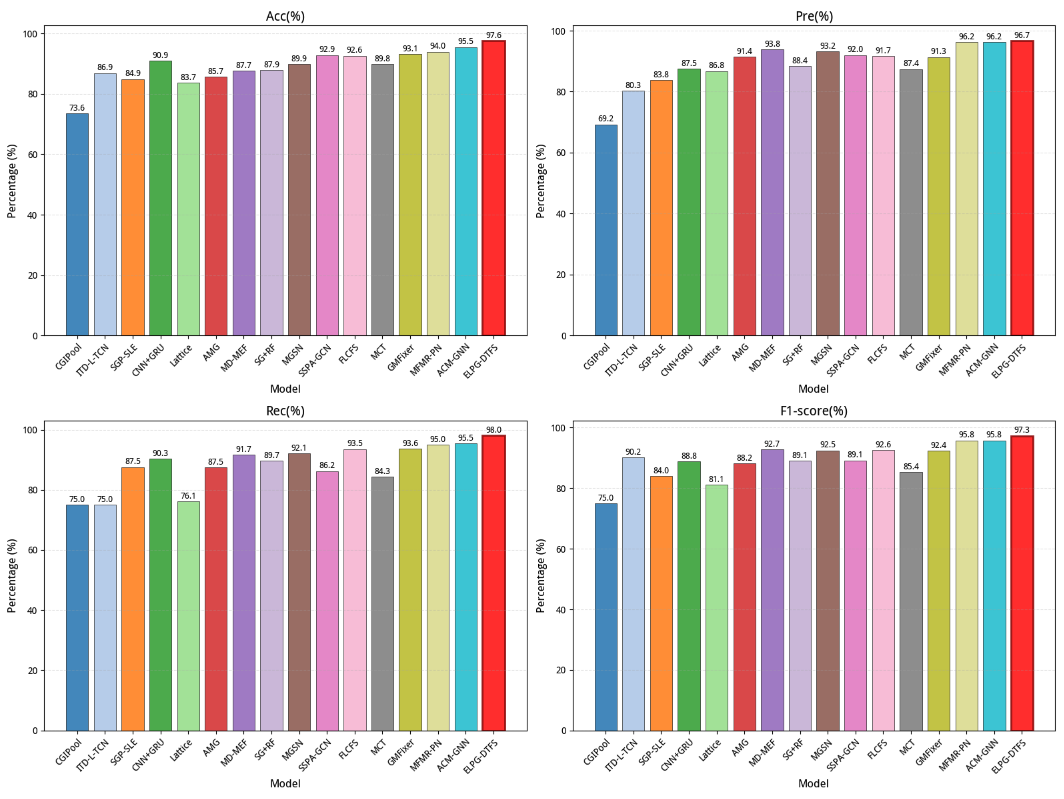}
    \caption{Mean performance and standard deviation over 10 folds on MODMA.}
    \label{fig:model_comparison}
\end{figure}

Table~\ref{tab:model_comparison} summarizes results for fourteen recent methods and ELPG-DTFS.  
Our model achieves \textbf{97.63 \%} Acc and \textbf{97.33 \%} F1, outperforming ACM-GNN by $+2.17$ and $+1.53$ percentage points ($p<0.01$).  
Figure~\ref{fig:model_comparison} visualizes the margin.

\begin{table}[t]
  \centering
  \caption{Performance comparison on MODMA (mean of 10 folds). Bold numbers indicate the best result; $^\S$ = metric not reported.}
  \label{tab:model_comparison}
  \setlength{\tabcolsep}{6pt}
  \renewcommand{\arraystretch}{1.1}
  {\small
  \begin{tabular}{lcccc}
    \toprule
    Model & Acc(\%) & Pre(\%) & Rec(\%) & F1(\%) \\
    \midrule
    CGIPool~\cite{ref18} (2021)     & 73.58 & 69.23 & 75.00    & 75.00 \\
    1TD+L--TCN~\cite{ref19} (2021)  & 86.87 & 83.83 & 90.15$^\S$ & 90.15 \\
    SGP--SLe~\cite{ref20} (2022)    & 84.91 & 80.77 & 87.54    & 84.00 \\
    CNN+GRU~\cite{ref21} (2022)     & 90.62 & 87.48 & 90.26    & 88.79 \\
    Lattice~\cite{ref22} (2023)     & 83.96 & 86.76 & 76.14    & 81.10 \\
    AMG~\cite{ref23} (2023)         & 88.68 & 91.43 & 87.50    & 88.17 \\
    MDD--MFF~\cite{ref24} (2024)    & 85.71 & 83.85 & 91.72    & 87.72 \\
    SG+RF~\cite{ref25} (2024)       & 87.87 & 88.42 & 89.71    & 89.06 \\
    MGSN~\cite{ref26} (2024)        & 89.87 & 93.16 & 86.07    & 89.47 \\
    SSPA--GCN~\cite{ref27} (2024)   & 92.87 & 92.00 & 92.24    & 92.12 \\
    FLFCFS~\cite{ref28} (2024)      & 92.59 & 91.67 & 93.55    & 92.60 \\
    MCT~\cite{ref29} (2025)         & 89.84 & 87.36 & 89.41    & 88.37 \\
    GNMixer~\cite{ref30} (2025)     & 93.12 & 84.29$^\S$ & 84.29 & 95.17 \\
    MFMR--FN~\cite{ref31} (2025)    & 93.96 & 94.95$^\S$ & 94.95 & 93.97 \\
    ACM--GNN~\cite{ref32} (2025)    & 95.46 & 96.23 & 95.46    & 95.80 \\
    \midrule
    \textbf{ELPG-DTFS}              & \textbf{97.63} & \textbf{96.68} & \textbf{98.03} & \textbf{97.33} \\
    \bottomrule
  \end{tabular}
  }
\end{table}

\textbf{Why does ELPG-DTFS outperform?}
{Temporal fidelity:} 
Baselines like CNN+GRU treat spectrograms as static images, blurring transient events that signal affective dysregulation. 
Our window-level channel-band attention preserves millisecond-scale details, which chiefly raises Recall to 98.03\%.
{Adaptive connectivity:} 
Fixed graphs in ACM-GNN omit weak yet meaningful couplings. 
The learnable mask $\mathbf{W}_{\alpha}$ reallocates weight to these edges, yielding a 2.17 pp accuracy gain.
{Hierarchical priors:} 
Regional priors steer the model toward established biomarkers—such as frontal alpha asymmetry—improving precision, especially under limited sample size.
{Cross-band coupling:} 
Mutual information captures nonlinear interactions (e.g., alpha–beta desynchronization) distinctive to MDD, refining decision boundaries and boosting F1.

\subsection{Ablation Study and Interpretation}\label{subsec:ablation}

Table~\ref{tab:ablation_study} shows degradations when modules are removed, and the following explains observed drops:

\begin{table}[h]
    \centering
    \caption{Ablation on MODMA (mean of 10 folds).}
    \label{tab:ablation_study}
    {\footnotesize
    \begin{tabular}{lcccc}
        \toprule
        Variant & Acc(\%) & Pre(\%) & Rec(\%) & F1(\%) \\
        \midrule
        Full ELPG-DTFS & \textbf{97.63} & \textbf{96.68} & \textbf{98.03} & \textbf{97.33} \\
        $-$ Prior knowledge    & 95.12 & 94.36 & 96.25 & 95.29 \\
        $-$ Learnable adjacency& 94.01 & 93.52 & 95.67 & 94.58 \\
        $-$ MI                 & 94.85 & 91.87 & 96.18 & 95.03 \\
        $-$ Attention \& MI    & 93.25 & 91.87 & 94.12 & 92.98 \\
        \bottomrule
    \end{tabular}
    }
\end{table}

{Prior knowledge}:  
    Both Precision and Recall fall, indicating that priors simultaneously reduce false positives and uncover true depressive cases by directing attention to neurobiologically grounded regions.{Learnable adjacency}:  
    Recall drops most, showing that adaptive edges principally help detect otherwise subtle pathological connectivity.{MI removal}:  
    Precision decreases, revealing MI’s role in filtering out healthy controls with benign spectral variations.{Attention \& MI removal}:  
    The greatest decline (4.35 pp F1) suggests dramatic loss of expressive power, leading to blurred class boundaries.

\vspace{-0.35em}

\section{Conclusion}
\label{sec:conclusion}
  
This paper presents ELPG-DTFS, a prior-guided adaptive graph neural network for EEG-based depression detection. It introduces channel–band attention for nonlinear cross-band dependencies, a learnable adjacency mask for dynamic functional graphs, and a multi-scale knowledge graph to incorporate clinical priors. On the MODMA benchmark, ELPG-DTFS achieves 97.63\% accuracy and 97.33\% F1, surpassing 14 recent baselines. Ablation confirms that all modules are indispensable, showing the value of combining signal processing, graph learning, and domain knowledge for objective depression diagnosis.


\end{document}